\documentclass[11pt]{article}
\usepackage{acl2015}
\usepackage{times}
\usepackage{url}
\usepackage{latexsym}

\usepackage{graphicx}
\usepackage{multirow}
\usepackage{amsmath,amsfonts,amssymb}

\title{Classifying Relations by Ranking with Convolutional Neural Networks}

\author{C\'{i}cero Nogueira dos Santos \\
  IBM Research \\
  138/146 Av. Pasteur \\
  Rio de Janeiro, RJ, Brazil \\
  {\tt cicerons@br.ibm.com} \\\And
  Bing Xiang \\
  IBM Watson \\
  1101 Kitchawan  \\
  Yorktown Heights, NY, USA \\
  {\tt bingxia@us.ibm.com} \\\And
  Bowen Zhou \\
  IBM Watson \\
  1101 Kitchawan \\
  Yorktown Heights, NY, USA \\
  {\tt zhou@us.ibm.com}
  \\}

\date{}

\begin{document}
\maketitle
\begin{abstract}
Relation classification is an important semantic processing task for which state-of-the-art systems still rely on costly handcrafted features.
In this work we tackle the relation classification task using a convolutional neural network that performs classification by ranking (CR-CNN).
We propose a new pairwise ranking loss function that makes it easy to reduce the impact of artificial classes.
We perform experiments using the the SemEval-2010 Task 8 dataset,
which is designed for the task of classifying the relationship between two nominals marked in a sentence.
Using CR-CNN,
we outperform the state-of-the-art for this dataset and achieve a F1 of 84.1 without using any costly handcrafted features.
Additionally, our experimental results show that: (1) our approach is more effective than CNN followed by a softmax classifier;
(2) omitting the representation of the artificial class \emph{Other} improves both precision and recall; and 
(3) using only word embeddings as input features is enough to achieve state-of-the-art results if we consider only the text between the two target nominals.
\end{abstract}


\section{Introduction}
\label{sec:intro}
Relation classification is an important Natural Language Processing (NLP) task which is normally used as an intermediate step in many complex NLP applications such as question-answering and automatic knowledge base construction.
Since the last decade there has been increasing interest in applying machine learning approaches to this task \cite{zhang2004:cikm,qian2009:SLS,rink:2010}.
One reason is the availability of benchmark datasets such as the SemEval-2010 task 8 dataset \cite{hendrickx2010:semeval},
which encodes the task of classifying the relationship between two nominals marked in a sentence. The following sentence contains an example of the \emph{Component-Whole} relation between the nominals \emph{``introduction''} and \emph{``book''}.

\begin{center}
\begin{small}
The [introduction]$_{e_1}$ in the [book]$_{e_2}$ is a \\
summary of what is in the text.
\end{small}
\end{center}

Some recent work on relation classification has focused on the use of deep neural networks with the aim of reducing the number of handcrafted features \cite{socher:2012:emnlp,zeng2014:coling,yu2014}. 
However,
in order to achieve state-of-the-art results these approaches still use some features derived from lexical resources such as WordNet or NLP tools such as dependency parsers and named entity recognizers (NER). 

In this work, 
we propose a new convolutional neural network (CNN), 
which we name Classification by Ranking CNN (CR-CNN), to tackle the relation classification task.
The proposed network learns a distributed vector representation for each relation class.
Given an input text segment,
the network uses a convolutional layer to produce a distributed vector representation of the text and compares it to the class representations in order to produce a score for each class.
We propose a new pairwise ranking loss function that makes it easy to reduce the impact of artificial classes.
We perform an extensive number of experiments using the the SemEval-2010 Task 8 dataset.
Using CR-CNN, and without the need for any costly handcrafted feature,
we outperform the state-of-the-art for this dataset.
Our experimental results are evidence that: 
(1) CR-CNN is more effective than CNN followed by a softmax classifier;
(2) omitting the representation of the artificial class \emph{Other} improves both precision and recall; and 
(3) using only word embeddings as input features is enough to achieve state-of-the-art results if we consider only the text between the two target nominals.

The remainder of the paper is structured as follows. Section~\ref{sec:neuralnetwork} details the proposed neural network. 
In Section~\ref{sec:experimental_setup}, 
we present details about the setup of experimental evaluation, 
and then describe the results in Section~\ref{sec:results}. 
In Section~\ref{sec:related_work}, 
we discuss previous work in deep neural networks for relation classification and for other NLP tasks. Section~\ref{sec:conclusion} presents our conclusions.
\section{The Proposed Neural Network}
\label{sec:neuralnetwork}
\begin{figure}[t]
  \centering
    \includegraphics[width=0.4\textwidth]{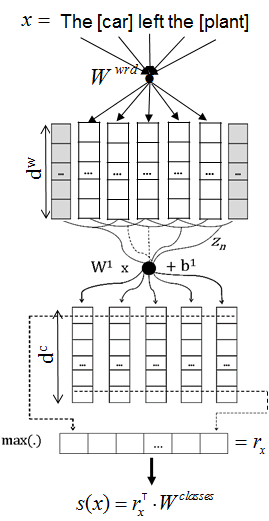}
  \caption{CR-CNN: a Neural Network for classifying by ranking.}
  \label{fig:cnn_pic}
\end{figure}

Given a sentence $x$ and two target nouns,
CR-CNN computes a score for each relation class $c \in C$.
For each class $c \in C$,
the network learns a distributed vector representation which is encoded as a column in the class embedding matrix $W^{classes}$.
As detailed in Figure~\ref{fig:cnn_pic},
the only input for the network is the tokenized text string of the sentence.
In the first step,
CR-CNN transforms words into real-valued feature vectors.
Next,
a convolutional layer is used to construct a distributed vector representations of the sentence, $r_{x}$.
Finally,
CR-CNN computes a score for each relation class $c \in C$ by performing a dot product between $r_{x}^\intercal$ and $W^{classes}$.

\subsection{Word Embeddings}
\label{sec:wordrepr}
The first layer of the network transforms words into representations that capture syntactic and semantic information about the words.
Given a sentence $x$ consisting of $N$ words $x=\{w_{1}, w_{2}, ..., w_{N}\}$, 
every word $w_{n}$ is converted into a real-valued vector $r^{w_{n}}$.
Therefore, 
the input to the next layer is a sequence of real-valued vectors $emb_{x}=\{r^{w_{1}}, r^{w_{2}}, ..., r^{w_{N}}\}$

Word representations are encoded by column vectors in an embedding matrix $W^{wrd}\in\mathbb{R}^{d^{w}\times|V|}$,
where $V$ is a fixed-sized vocabulary.
Each column $W^{wrd}_{i}\in\mathbb{R}^{d^{w}}$ corresponds to the word embedding of the $i$-th word in the vocabulary.
We transform a word $w$ into its word embedding $r^{w}$ by using the matrix-vector product: 
\begin{displaymath} \label{word_representation}
r^{w} = W^{wrd}v^{w}
\end{displaymath}
where $v^{w}$ is a vector of size $\left|V\right|$ which has value $1$ at index $w$ and zero in all other positions.
The matrix $W^{wrd}$ is a parameter to be learned,
and the size of the word embedding $d^{w}$ is a hyperparameter to be chosen by the user.

\subsection{Word Position Embeddings}
In the task of relation classification,
information that is needed to determine the class of a relation between two target nouns normally comes from words which are close to the target nouns.
Zeng et al. \shortcite{zeng2014:coling} propose the use of \emph{word position embeddings} (position features) which help the CNN by keeping track of how close words are to the target nouns.
These features are similar to the position features proposed by Collobert et al. \shortcite{collobert2011:jmlr} for the Semantic Role Labeling task.

In this work we also experiment with the word position embeddings (WPE) proposed by Zeng et al. \shortcite{zeng2014:coling}.
The WPE is derived from the relative distances of the current word to the target $noun_1$ and $noun_2$. 
For instance, 
in the sentence shown in Figure \ref{fig:cnn_pic},
the relative distances of \emph{“left”} to \emph{“car”} and \emph{“plant”} are -1
and 2, 
respectively. 
As in \cite{collobert2011:jmlr},
each relative distance is mapped to a vector of dimension $d^{wpe}$,
which is initialized with random numbers.
$d^{wpe}$ is a hyperparameter of the network.
Given the vectors $wp_1$ and $wp_2$ for the word $w$ with respect to the targets $noun_1$ and $noun_2$,
the position embedding of $w$ is given by the concatenation of these two vectors, $wpe^{w}=[wp_1,wp_2]$.

In the experiments where word position embeddings are used,
the word embedding and the word position embedding of each word are concatenated to form the input for the convolutional layer, $emb_{x}=\{[r^{w_{1}},wpe^{w_1}], [r^{w_{2}},wpe^{w_2}], ..., [r^{w_{N}},wpe^{w_N}]\}$.

\subsection{Sentence Representation}
The next step in the NN consists in creating the distributed vector representation $r_{x}$ for the input sentence $x$.
The main challenges in this step are the sentence size variability and the fact that important information can appear at any position in the sentence.
In recent work,
convolutional approaches have been used to tackle these issues when creating representations for text segments of different sizes \cite{zeng2014:coling,Hu@NIPS2014,santos2014:coling} and
character-level representations of words of different sizes \cite{santos2014:icml}.
Here,
we use a convolutional layer to compute distributed vector representations of the sentence.
The convolutional layer first produces local features around each word in the sentence.
Then, 
it combines these local features using a max operation to
create a fixed-sized vector for the input sentence.
 
Given a sentence $x$,
the convolutional layer applies a matrix-vector operation to each window of size $k$ of successive windows in $emb_{x}=\{r^{w_{1}}, r^{w_{2}}, ..., r^{w_{N}}\}$.
Let us define the vector $z_n\in\mathbb{R}^{d^{w}k}$ as the concatenation of a sequence of $k$ word embeddings,
centralized in the $n$-th word:
\begin{displaymath}
z_n=\left(r^{w_{n-(k-1)/2}}, ..., r^{w_{n+(k-1)/2}}\right)^\intercal
\end{displaymath}
In order to overcome the issue of referencing words with indices outside of the sentence boundaries, 
we augment the sentence with a special \emph{padding} token replicated $\dfrac{k-1}{2}$ times at the beginning and the end.

The convolutional layer computes the $j$-th element of the vector $r_{x} \in \mathbb{R}^{d^{c}}$ as follows:
\begin{displaymath}
[r_{x}]_j = \max_{1 < n < N}\left[f\left(W^{1}z_n + b^{1}\right)\right]_j
\end{displaymath}
where $W^{1} \in \mathbb{R}^{d^{c} \times d^{w}k}$ is the weight matrix of the convolutional layer
and $f$ is the hyperbolic tangent function.
The same matrix is used to extract local features around each word window of the given sentence.
The fixed-sized distributed vector representation for the sentence is obtained by using the $max$ over all word windows.
Matrix $W^{1}$ and vector $b^{1}$ are parameters to be learned.
The number of convolutional units $d^{c}$,
and the size of the word context window $k$
are hyperparameters to be chosen by the user.
It is important to note that $d^{c}$ corresponds to the size of the sentence representation.

\subsection{Class embeddings and Scoring}
Given the distributed vector representation of the input sentence $x$, 
the network with parameter set $\theta$ computes the score for a class label $c \in C$ by using the dot product
\begin{displaymath}
s_{\theta}(x)_c=r_{x}^\intercal [W^{classes}]_c
\end{displaymath}
where $W^{classes}$ is an embedding matrix whose columns encode the distributed vector representations of the different class labels, 
and $[W^{classes}]_c$ is the column vector that contains the embedding of the class $c$.
Note that the number of dimensions in each class embedding must be equal to the size of the sentence representation,
which is defined by $d^{c}$.
The embedding matrix $W^{classes}$ is a parameter to be learned by the network.
It is initialized by randomly sampling each value from an uniform distribution: $\mathcal{U}\left(-r,r\right)$,
where $r=\sqrt{\dfrac{6}{|C|+d^{c}}}$.

\subsection{Training Procedure}
Our network is trained by minimizing a pairwise ranking loss function over the training set $D$.
The input for each training round is a sentence $x$ and two different class labels $y^+ \in C$ and $c^- \in C$,
where $y^+$ is a correct class label for $x$ and $c^-$ is not.
Let $s_{\theta}(x)_{y^+}$ and $s_{\theta}(x)_{c^-}$ be respectively the scores for class labels $y^+$ and $c^-$ generated by the network with parameter set $\theta$.
We propose a new logistic loss function over these scores in order to train CR-CNN:
\begin{equation}
\arraycolsep=1.4pt\def\arraystretch{1.5}
\begin{array}{ll}
L = & \ \ log(1+exp(\gamma (m^+ - s_{\theta}(x)_{y^+})) \\ 
                 & + \ log(1+exp(\gamma (m^- + s_{\theta}(x)_{c^-}))\\
\end{array}
\label{eq:lossfunction}
\end{equation}
where $m^+$ and $m^-$ are margins and $\gamma$ is a scaling factor that magnifies the difference between the score and the margin and helps to penalize more on the prediction errors.
The first term in the right side of Equation \ref{eq:lossfunction} decreases as the score $s_{\theta}(x)_{y^+}$ increases.
The second term in the right side decreases as the score $s_{\theta}(x)_{c^-}$ decreases.
Training CR-CNN by minimizing the loss function in Equation \ref{eq:lossfunction} has the effect of training to give scores greater than $m^+$ for the correct class and (negative) scores smaller than $-m^-$ for incorrect classes.
In our experiments we set $\gamma$ to 2,
$m^+$ to 2.5 and $m^-$ to 0.5.
We use $L2$ regularization by adding the term $\beta\|\theta\|^2$ to Equation \ref{eq:lossfunction}.
In our experiments we set $\beta$ to 0.001.
We use stochastic gradient descent (SGD) to minimize the
loss function with respect to $\theta$.

Like some other ranking approaches that only update two classes/examples at every training round \cite{weston:ijcai2011,gao2014:emnlp},
we can efficiently train the network for tasks which have a very large number of classes.
This is an advantage over softmax classifiers.

On the other hand,
sampling informative negative classes/examples can have a significant impact in the effectiveness of the learned model.
In the case of our loss function,
more informative negative classes are the ones with a score larger than $-m^-$.
The number of classes in the relation classification dataset that we use in our experiments is small.
Therefore,
in our experiments,
given a sentence $x$ with class label $y^+$,
the incorrect class $c^-$ that we choose to perform a SGD step is the one with the highest score among all incorrect classes  $c^-=\displaystyle\operatorname*{arg\,max}_{c \ \in \ C; \ c \neq  y^+} s_{\theta}(x)_{c}$.

For tasks where the number of classes is large,
we can fix a number of negative classes to be considered at each example and select the one with the largest score to perform a gradient step.
This approach is similar to the one used by Weston et al. \shortcite{weston:emnlp2014} to select negative examples.

We use the backpropagation algorithm to compute gradients of the network.
In our experiments, 
we implement the CR-CNN architecture and the backpropagation algorithm using Theano \cite{bergstra:scipy2010}.

\subsection{Special Treatment of \emph{Artificial} Classes}
\label{sec:omitting_classes}
In this work,
we consider a class as \emph{artificial} if it is used to group items that do not belong to any of the \emph{actual} classes.
An example of artificial class is the class \emph{Other} in the SemEval 2010 relation classification task.
In this task,
the artificial class \emph{Other} is used to indicate that the relation between two nominals does not belong to any of the nine relation classes of interest.
Therefore,
the class \emph{Other} is very noisy since it groups many different types of relations that may not have much in common. 

An important characteristic of CR-CNN is that it makes it easy to reduce the effect of artificial classes by omitting their embeddings.
If the embedding of a class label $c$ is omitted,
it means that the embedding matrix $W^{classes}$ does not contain a column vector for $c$.
One of the main benefits from this strategy is that the learning process focuses on the ``natural'' classes only. Since the embedding of the artificial class is omitted, 
it will not influence the prediction step, 
i.e., 
CR-CNN does not produce a score for the artificial class.

In our experiments with the SemEval-2010 relation classification task,
when training with a sentence $x$ whose class label $y=Other$,
the first term in the right side of Equation \ref{eq:lossfunction} is set to zero.
During prediction time,
a relation is classified as \emph{Other} only if all actual classes have negative scores.
Otherwise, 
it is classified with the class which has the largest score.

\section{Experimental Setup}
\label{sec:experimental_setup}

\subsection{Dataset and Evaluation Metric}
We use the SemEval-2010 Task 8 dataset to perform our experiments.
This dataset contains 10,717 examples annotated with 9 different relation types and an artificial relation Other,
which is used to indicate that the relation in the example does not belong to any of the nine main relation types.
The nine relations are \emph{Cause-Effect, 
Component-Whole, 
Content-Container,
Entity-Destination,
Entity-Origin,
Instrument-Agency,
Member-Collection,
Message-Topic}
and \emph{Product-Producer}.
Each example contains a sentence marked with two nominals $e_1$ and $e_2$,
and the task consists of predicting the relation between the two nominals taking into consideration the directionality.
That means that the relation Cause-Effect(e1,e2) is different from the relation Cause-Effect(e2,e1), as shown in the examples below.
More information about this dataset can be found in \cite{hendrickx2010:semeval}.

\begin{center}
\begin{small}
The [war]$_{e_1}$ resulted in other collateral imperial [conquests]$_{e_2}$ as well. $\Rightarrow$ Cause-Effect(e1,e2)

The [burst]$_{e_1}$ has been caused by water hammer [pressure]$_{e2}$. $\Rightarrow$ Cause-Effect(e2,e1)
\end{small}
\end{center}

The SemEval-2010 Task 8 dataset is already partitioned into 8,000 training instances and 2,717 test instances.
We score our systems by using the SemEval-2010 Task 8 official scorer, which computes the macro-averaged F1-scores for the nine actual relations (excluding Other) and takes the directionality into consideration.

\subsection{Word Embeddings Initialization}
The word embeddings used in our experiments are initialized by means of unsupervised pre-training.
We perform pre-training using the skip-gram NN architecture~\cite{mikolov:2013} available in the \texttt{word2vec} tool.
We use the December 2013 snapshot of the English Wikipedia corpus to train word embeddings with \texttt{word2vec}.
We preprocess the Wikipedia text using the steps described in \cite{santos2014:coling}:
(1) removal of paragraphs that are not in English;
(2) substitution of non-western characters for a special character;
(3) tokenization of the text using the tokenizer available with the Stanford POS Tagger \cite{toutanova2003:naacl};
(4) removal of sentences that are less than 20 characters long (including white spaces) or have less than 5 tokens.
(5) lowercase all words and substitute each numerical digit by a 0.
The resulting clean corpus contains about 1.75 billion tokens.

\subsection{Neural Network Hyper-parameter}
We use 4-fold cross-validation to tune the neural network hyperparameters.
Learning rates in the range of 0.03 and 0.01 give relatively similar results.
Best results are achieved using between 10 and 15 training epochs,
depending on the CR-CNN configuration.
In Table \ref{tab:nn_hyperparams}, 
we show the selected hyperparameter values.
Additionally,
we use a learning rate schedule that decreases the learning rate $\lambda$ according to the training epoch $t$.
The learning rate for epoch $t$, 
$\lambda_t$, 
is computed using the equation:
$\lambda_t = \dfrac{\lambda}{t}$.

\begin{table}[h!]
\begin{center}
\begin{tabular}{|l|l|r|}
\hline  
\bf Parameter & \bf Parameter Name & \bf Value\\
\hline
$d^{w}$ & Word Emb. size          & 400   \\
$d^{wpe}$ & Word Pos. Emb. size   & 70   \\
$d^{c}$   & Convolutinal Units    & 1000  \\ 
$k$       & Context Window size   & 3     \\ 
$\lambda$ & Initial Learning Rate & 0.025 \\
\hline
\end{tabular}
\end{center}
\caption{\label{tab:nn_hyperparams} CR-CNN Hyperparameters }
\end{table}

\section{Experimental Results}
\label{sec:results}

\subsection{Word Position Embeddings and Input Text Span}
\label{sec:textspan}
In the experiments discussed in this section we assess the impact of using word position embeddings (WPE) and also propose a simpler alternative approach that is almost as effective as WPEs. 
The main idea behind the use of WPEs in relation classification task is to give some hint to the convolutional layer of how close a word is to the target nouns,
based on the assumption that closer words have more impact than distant words.

Here we hypothesize that most of the information needed to classify the relation appear between the two target nouns.
Based on this hypothesis, 
we perform an experiment where the input for the convolutional layer consists of the word embeddings of the word sequence $\{w_{e_1}-1,...,w_{e_2}+1\}$ where $e_1$ and $e_2$ correspond to the positions of the first and the second target nouns, respectively. 

In Table \ref{tab:res:text_span} we compare the results of different CR-CNN configurations.
The first column indicates whether the full sentence was used (\emph{Yes}) or whether the text span between the target nouns was used (\emph{No}). 
The second column informs if the WPEs were used or not.
It is clear that the use of WPEs is essential when the full sentence is used,
since F1 jumps from 74.3 to 84.1.
This effect of WPEs is reported by \cite{zeng2014:coling}.
On the other hand,
when using only the text span between the target nouns,
the impact of WPE is much smaller.
With this strategy,
we achieve a F1 of 82.8 using only word embeddings as input, 
which is a result as good as the previous state-of-the-art F1 of 83.0 reported in \cite{yu2014} for the SemEval-2010 Task 8 dataset.
This experimental result also suggests that,
in this task,
the CNN works better for short texts.

All experiments reported in the next sections use CR-CNN with full sentence and WPEs.

\begin{table}[h!]
\begin{center}
\begin{tabular}{|c|c|ccc|}
\hline  \bf Full  & \bf Word & \multirow{2}{*}{\bf Prec.} & \multirow{2}{*}{\bf Rec.} & \multirow{2}{*}{\bf F1} \\ 
     \bf Sentence & \bf Position & & & \\
\hline
Yes & Yes & \bf 83.7 & \bf 84.7 & \bf 84.1 \\
No  & Yes & 83.3 & 83.9 & 83.5 \\
No  & No  & 83.4 & 82.3 & 82.8 \\
Yes & No  & 78.1 & 71.5 & 74.3\\
\hline
\end{tabular}
\end{center}
\caption{\label{tab:res:text_span} Comparison of different CR-CNN configurations. }
\end{table}

\subsection{Impact of Omitting the Embedding of the artificial class \emph{Other}}
In this experiment we assess the impact of omitting the embedding of the class \emph{Other}.
As we mentioned above,
this class is very noisy since it groups many different infrequent relation types.
Its embedding is difficult to define and therefore brings noise into the classification process of the natural classes.
In Table \ref{tab:res:classother} we present the results comparing the use and omission of embedding for the class \emph{Other}.
The two first lines of results present the official F1,
which does not take into account the results for the class \emph{Other}.
We can see that by omitting the embedding of the class \emph{Other} both precision and recall for the other classes improve,
which results in an increase of 1.4 in the F1.
These results suggest that the strategy we use in CR-CNN to avoid the noise of artificial classes is effective.

\begin{table}[h!]
\begin{center}
\begin{tabular}{|c|c|ccc|}
\hline  \bf Use embedding & \multirow{2}{*}{\bf Class} & \multirow{2}{*}{\bf Prec.} & \multirow{2}{*}{\bf Rec.} & \multirow{2}{*}{\bf F1} \\ 
     \bf of class \emph{Other} & & & & \\
\hline
No  & All   & \bf 83.7 & \bf 84.7 & \bf 84.1 \\
Yes & All   &  81.3    & 84.3     &  82.7   \\
\hline
No  & Other & 52.0 & 48.7 & 50.3 \\
Yes & Other & 60.1 & 48.7 & 53.8 \\
\hline
\end{tabular}
\end{center}
\caption{\label{tab:res:classother} Impact of not using an embedding for the artificial class Other. }
\end{table}

In the two last lines of Table \ref{tab:res:classother} we present the results for the class \emph{Other}.
We can note that while the recall for the cases classified as \emph{Other} remains 48.7,
the precision significantly decreases from 60.1 to 52.0 when the embedding of the class \emph{Other} is not used.
That means that more cases from natural classes (all) are now been classified as \emph{Other}.
However,
as both the precision and the recall of the natural classes increase,
the cases that are now classified as \emph{Other} must be cases that are also wrongly classified when the embedding of the class \emph{Other} is used.



\subsection{CR-CNN versus CNN+Softmax}
In this section we report experimental results comparing CR-CNN with CNN+Softmax.
In order to do a fair comparison,
we've implemented a CNN+Softmax and trained it with the same data, word embeddings and WPEs used in CR-CNN.
Concretely, 
our CNN+Softmax consists in getting the output of the convolutional layer,
which is the vector $r_x$ in Figure~\ref{fig:cnn_pic},
and giving it as input for a softmax classifier.
We tune the parameters of CNN+Softmax by using a 4-fold cross-validation with the training set.
Compared to the hyperparameter values for CR-CNN presented in Table \ref{tab:nn_hyperparams},
the only difference for CNN+Softmax is the number of convolutional units $d^{c}$, 
which is set to 400.

In Table \ref{tab:res:cnnsoftmax} we compare the results of CR-CNN and CNN+Softmax.
CR-CNN outperforms CNN+Softmax in both precision and recall,
and improves the F1 by 1.6.
The third line in Table \ref{tab:res:cnnsoftmax} shows the result reported by Zeng et al. \shortcite{zeng2014:coling} when only word embeddings and WPEs are used as input to the network (similar to our CNN+Softmax).
We believe that the word embeddings employed by them is the main reason their result is much worse than that of CNN+Softmax.
We use word embeddings of size 400 while they use word embeddings of size 50,
which were trained using much less unlabeled data than we did. 

\begin{table}[h!]
\begin{center}
\begin{tabular}{|c|ccc|}
\hline  \bf Neural Net. & \bf \bf Prec. & \bf Rec. & \bf F1 \\ 
\hline
CR-CNN      & \bf 83.7 & \bf 84.7 & \bf 84.1 \\
CNN+SoftMax & 82.1     & 83.1     & 82.5 \\
\hline
CNN+SoftMax  & \multirow{2}{*}{-} & \multirow{2}{*}{-} & \multirow{2}{*}{78.9} \\
\cite{zeng2014:coling} &  & & \\ 
\hline
\end{tabular}
\end{center}
\caption{\label{tab:res:cnnsoftmax} Comparison of results of CR-CNN and CNN+Softmax.}
\end{table}

\subsection{Comparison with the State-of-the-art}
In Table \ref{tab:res:state_of_the_art} we compare CR-CNN results with results recently published for the SemEval-2010 Task 8 dataset.
Rink and Harabagiu \shortcite{rink:2010} present a support vector machine (SVM) classifier that is fed with a rich (traditional) feature set. It obtains an F1 of 82.2, 
which was the best result at SemEval-2010 Task 8. 
Socher et al. \shortcite{socher:2012:emnlp} present results for a recursive neural network (RNN) that employs a matrix-vector representation to every node in a parse tree in order to compose the distributed vector representation for the complete sentence.
Their method is named the matrix-vector recursive neural network (MVRNN)
and achieves a F1 of 82.4 when POS, NER and WordNet features are used.
In \cite{zeng2014:coling},
the authors present results for a CNN+Softmax classifier which employs lexical and sentence-level features.
Their classifier achieves a F1 of 82.7 when adding a handcrafted feature based on the WordNet. 
Yu et al. \shortcite{yu2014} present the Factor-based Compositional Embedding Model (FCM),
which achieves a F1 of 83.0 by deriving sentence-level and substructure embeddings from word embeddings utilizing dependency trees and named entities.

As we can see in the last line of Table \ref{tab:res:state_of_the_art},
CR-CNN using the full sentence, word embeddings and WPEs outperforms all previous reported results and reaches a new state-of-the-art F1 of 84.1.
This is a remarkable result since we do not use any complicated features that depend on external lexical resources such as WordNet and NLP tools such as named entity recognizers (NERs) and dependency parsers.

We can see in Table \ref{tab:res:state_of_the_art} that CR-CNN\footnote{This is the result using only the text span between the target nouns.} also achieves the best result among the systems that use word embeddings as the only input features.
The closest result (80.6),
which is produced by the FCM system of Yu et al. \shortcite{yu2014},
is 2.2 F1 points behind CR-CNN result (82.8).

\begin{table*}[ht!]
\begin{small}
\begin{center}
\begin{tabular}{|c|l|c|}
\hline  \bf Classifier & \bf Feature Set & \bf F1 \\
\hline
SVM  & POS, prefixes, morphological, WordNet, dependency parse,  & \multirow{3}{*}{82.2}\\
\cite{rink:2010} & Levin classes, ProBank, FrameNet, NomLex-Plus, &  \\
& Google n-gram, paraphrases, TextRunner  & \\
\hline
RNN & word embeddings & 74.8 \\
\cite{socher:2012:emnlp} & word embeddings, POS, NER, WordNet & 77.6 \\
\hline
MVRNN &  word embeddings & 79.1 \\
\cite{socher:2012:emnlp} & word embeddings, POS, NER, WordNet & 82.4 \\
\hline
             & word embeddings & 69.7 \\
CNN+Softmax  & word embeddings, word position embeddings, & \multirow{2}{*}{82.7} \\
\cite{zeng2014:coling} & word pair, words around word pair, WordNet & \\
\hline
FCM & word embeddings & 80.6 \\
\cite{yu2014} & word embeddings, dependency parse, NER & 83.0 \\
\hline
\multirow{2}{*}{CR-CNN} & word embeddings & 82.8 \\
    & word embeddings, word position embeddings & \bf 84.1 \\
\hline
\end{tabular}
\end{center}
\end{small}
\caption{\label{tab:res:state_of_the_art} Comparison with results published in the literature.}
\end{table*}

\subsection{Most Representative Trigrams for each Relation}

\begin{table*}[ht!]
\begin{small}
\begin{center}
\begin{tabular}{|l|c|c|}
\hline  
  \bf Relation & \bf (e1,e2) & \bf (e2,e1)  \\
\hline
 \multirow{2}{*}{Cause-Effect} & $e1$ resulted in, $e1$ caused a, had caused & $e2$ caused by, was caused by, are \\
                               & the, poverty cause $e2$, caused a $e2$  & caused by, been caused by, $e2$ from $e1$ \\
\hline
  \multirow{2}{*}{Component-Whole} & $e1$ of the, of the $e2$, part of the, & $e2$ 's $e1$, with its $e1$, $e2$ has a, \\
                                   & in the $e2$, $e1$ on the  &  $e2$ comprises the, $e2$ with $e1$ \\
\hline
 \multirow{2}{*}{Content-Container} & was in a, was hidden in, were in a, & $e2$ full of, $e2$ with $e1$, $e2$ was full, \\
                    & was inside a, was contained in      & $e2$ contained a, $e2$ with cold \\
\hline
 \multirow{2}{*}{Entity-Destination} & $e1$ into the, $e1$ into a, $e1$ to the, & \multirow{2}{*}{-}\\
                    & was put inside, imported into the  & \\
\hline
 \multirow{2}{*}{Entity-Origin} & away from the, derived from a, had & the source of, $e2$ grape $e1$, \\
                      & left the, derived from an, $e1$ from the       &  $e2$ butter $e1$ \\
\hline
 \multirow{2}{*}{Instrument-Agency} & are used by, $e1$ for $e2$, is used by, & with a $e1$, by using $e1$, $e2$ finds a, \\
                                    & trade for $e2$, with the $e2$   & $e2$ with a, $e2$ \emph{,} who \\
\hline
 \multirow{2}{*}{Member-Collection} & of the $e2$, in the $e2$, of this $e2$,  & $e2$ of $e1$, of wild $e1$, of elven $e1$, \\
                                    & the political $e2$, $e1$ collected in  &  $e2$ of different, of 0000 $e1$ \\
\hline
 \multirow{3}{*}{Message-Topic}     & $e1$ is the, $e1$ asserts the, $e1$ that the, & described in the, discussed in the, \\
                                    & on the $e2$, $e1$ inform about  & featured in numerous, discussed  \\
                                    &                             & in cabinet, documented in two, \\
\hline
 \multirow{2}{*}{Product-Producer}  & $e1$ by the, by a $e2$, of the $e2$, & $e2$ of the, $e2$ has constructed, $e2$ 's $e1$, \\
                                    & by the $e2$, from the $e2$         &  $e2$ came up, $e2$ who created \\
\hline
\end{tabular}
\end{center}
\end{small}
\caption{\label{tab:res:ngrams} List of most representative trigrams for each relation type.}
\end{table*}

In Table \ref{tab:res:ngrams},
for each relation type we present the five trigrams in the test set which contributed the most for scoring correctly classified examples.
Remember that in CR-CNN,
given a sentence $x$ the score for the class $c$ is computed by $s_{\theta}(x)_c=r_{x}^\intercal [W^{classes}]_c$.
In order to compute the most representative trigram of a sentence $x$,
we trace back each position in $r_{x}$ to find the trigram responsible for it.
For each trigram $t$,
we compute its particular contribution for the score by summing the terms in score that use positions in $r_{x}$ that trace back to $t$.
The most \emph{representative} trigram in $x$ is the one with the largest contribution to the improvement of the score.
In order to create the results presented in Table \ref{tab:res:ngrams},
we rank the trigrams which were selected as the most representative of any sentence in decreasing order of contribution value.
If a trigram appears as the largest contributor for more than one sentence, 
its contribuition value becomes the sum of its contribution for each sentence.

We can see in Table \ref{tab:res:ngrams} that for most classes,
the trigrams that contributed the most to increase the score are indeed very informative regarding the relation type.
As expected,
different trigrams play an important role depending on the direction of the relation.
For instance,
the most informative trigram for \emph{Entity-Origin(e1,e2)} is \emph{``away from the''},
while reverse direction of the relation, \emph{Entity-Origin(e2,e1)} or \emph{Origin-Entity}, has \emph{``the source of''} as the most informative trigram.
These results are a step towards the extraction of meaningful knowledge from models produced by CNNs. 

\section{Related Work}
\label{sec:related_work}

Over the years, various approaches have been proposed for relation classification \cite{zhang2004:cikm,qian2009:SLS,hendrickx2010:semeval,rink:2010}. Most of them treat it as a multi-class classification problem and apply a variety of machine learning techniques to the task in order to achieve a high accuracy. 

Recently, 
deep learning \cite{bengio:2009} has become an attractive area for multiple applications, 
including computer vision, 
speech recognition and 
natural language processing.
Among the different deep learning strategies,
convolutional neural networks have been successfully applied to different NLP task such as part-of-speech tagging \cite{santos2014:icml},
sentiment analysis \cite{Kim@EMNLP2014,santos2014:coling},
question classification \cite{Kalchbrenner:acl2014},
semantic role labeling \cite{collobert2011:jmlr},
hashtag prediction \cite{weston:emnlp2014},
sentence completion and response matching \cite{Hu@NIPS2014}.

Some recent work on deep learning for relation classification include Socher et al. \shortcite{socher:2012:emnlp},
Zeng et al. \shortcite{zeng2014:coling}
and Yu et al. \shortcite{yu2014}.
In \cite{socher:2012:emnlp},
the authors tackle relation classification using a recursive neural network (RNN) that assigns a matrix-vector representation to every node in a parse tree.
The representation for the complete sentence is computed bottom-up by recursively combining the words according to the syntactic structure of the parse tree
Their method is named the matrix-vector recursive neural network (MVRNN).

Zeng et al. \shortcite{zeng2014:coling} propose an approach for relation classification where sentence-level features are learned through a CNN, 
which has word embedding and position features as its input. In parallel, lexical features are extracted according to given nouns. 
Then sentence-level and lexical features are concatenated into a single vector and fed into a softmax classifier for prediction. 
This approach achieves state-of-the-art performance on the SemEval-2010 Task 8 dataset.

Yu et al. \shortcite{yu2014} propose a Factor-based Compositional Embedding Model (FCM) by deriving sentence-level and substructure embeddings from word embeddings, utilizing dependency trees and named entities. It achieves slightly higher accuracy on the same dataset than \cite{zeng2014:coling}, but only when syntactic information is used.

There are two main differences between the approach proposed in this paper and the ones proposed in \cite{socher:2012:emnlp,zeng2014:coling,yu2014}:
(1) CR-CNN uses a pair-wise ranking method, while other approaches apply multi-class classification by using the softmax function on the top of the CNN/RNN;
and (2) CR-CNN employs an effective method to deal with artificial classes by omitting their embeddings, 
while other approaches treat all classes equally.

\section{Conclusion}
\label{sec:conclusion}
In this work we tackle the relation classification task using a CNN that performs classification by ranking.
The main contributions of this work are:
(1) the definition of a new state-of-the-art for the SemEval-2010 Task 8 dataset without using any costly handcrafted features;
(2) the proposal of a new CNN for classification that uses class embeddings and a new rank loss function;
(3) an effective method to deal with artificial classes by omitting their embeddings in CR-CNN;
(4) the demonstration that using only the text between target nominals is almost as effective as using WPEs; and
(5) a method to extract from the CR-CNN model the most representative contexts of each relation type.
Although we apply CR-CNN to relation classification, this method can be used for any classification task.


\section*{Acknowledgments}
The authors would like to thank Nina Wacholder for her  valuable suggestions to improve the final version of the paper.

\bibliographystyle{acl}
\bibliography{relation_classification_acl2015}

\end{document}